\documentclass[10pt,twocolumn,letterpaper]{article}

\usepackage{cvpr}              % To produce the CAMERA-READY version
\usepackage[dvipsnames]{xcolor}

\definecolor{cvprblue}{rgb}{0.21,0.49,0.74}
\usepackage[pagebackref,breaklinks,colorlinks,citecolor=cvprblue]{hyperref}

\def\paperID{12436} % *** Enter the Paper ID here
\def\confName{CVPR}
\def\confYear{2024}

\title{AttriHuman-3D: Editable 3D Human Avatar Generation with Attribute Decomposition and Indexing}

\author{ Fan Yang$^1$  \quad Tianyi Chen$^2$\quad   Xiaosheng He$^1$ \quad Zhongang Cai$^{1,3}$ \\ Lei Yang$^3$\quad  Si Wu$^2$ \quad  Guosheng Lin$^{1}$\thanks{Guosheng Lin is the corresponding author.}\\
$^1$S-Lab, Nanyang Technological University \\ $^2$School of Computer Science and Engineering, South China University of Technology \\ $^3$SenseTime Research \\ {\tt\small fan007@e.ntu.edu.sg}, {\tt\small csttychen@mail.scut.edu.cn}, {\tt\small gslin@ntu.edu.sg}}

\begin{document}
\maketitle
\begin{abstract}
Editable 3D-aware generation, which supports user-interacted editing, has witnessed rapid development recently. However, existing editable 3D GANs either fail to achieve high-accuracy local editing or suffer from huge computational costs. We propose AttriHuman-3D, an editable 3D human generation model, which address the aforementioned problems with attribute decomposition and indexing. The core idea of the proposed model is to generate all attributes (e.g. human body, hair, clothes and so on) in an overall attribute space with six feature planes, which are then decomposed and manipulated with different attribute indexes. To precisely extract features of different attributes from the generated feature planes, we propose a novel attribute indexing method as well as an orthogonal projection regularization to enhance the disentanglement. We also introduce a hyper-latent training strategy and an attribute-specific sampling strategy to avoid style entanglement and misleading punishment from the discriminator. Our method allows users to interactively edit selected attributes in the generated 3D human avatars while keeping others fixed. Both qualitative and quantitative experiments demonstrate that our model provides a strong disentanglement between different attributes, allows fine-grained image editing and generates high-quality 3D human avatars. 
\end{abstract}    
\section{Introduction}
\label{sec:intro}

\begin{figure*}
  \centering
    \includegraphics[width=\linewidth]{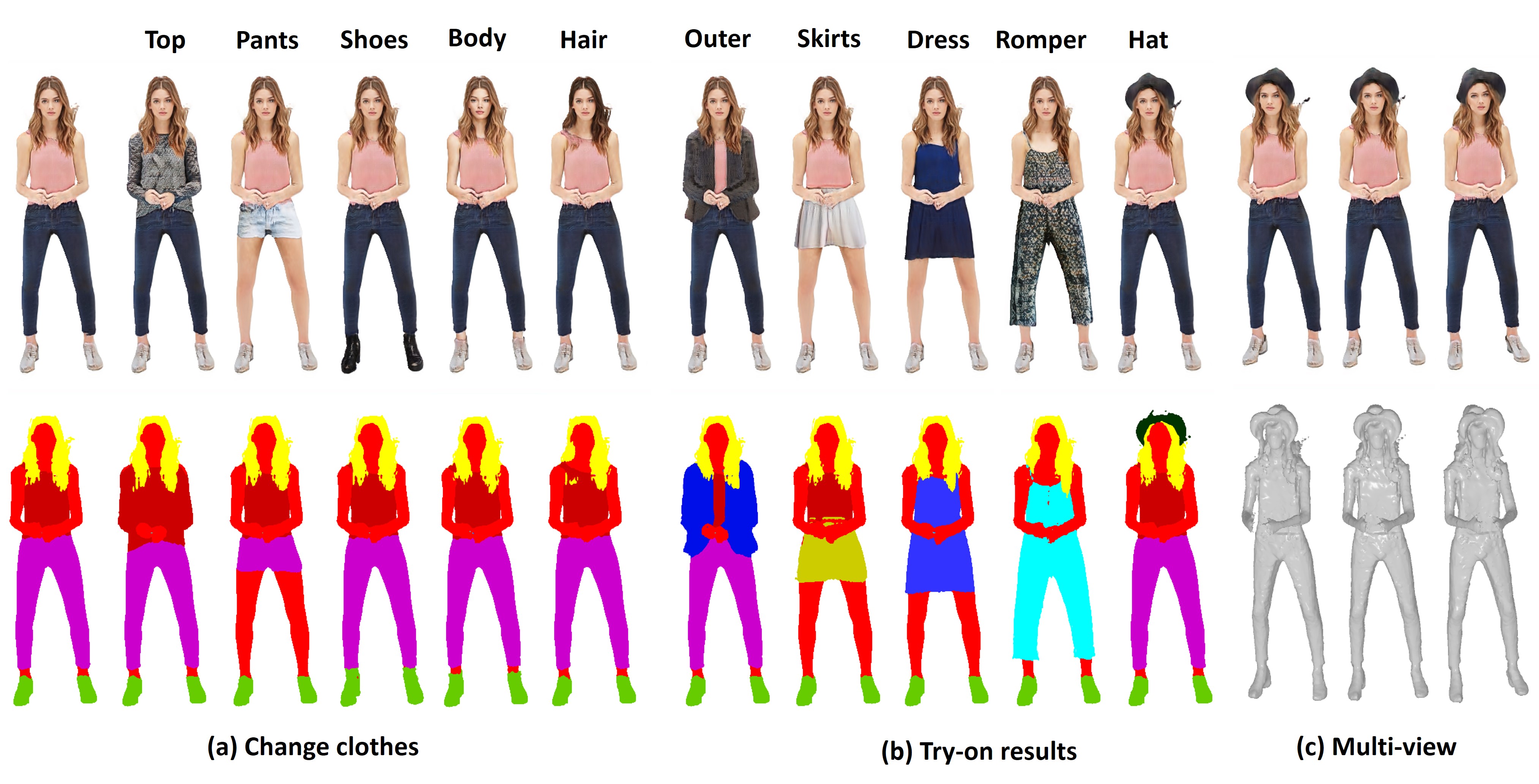}
    \label{pipeline}
  \caption{Our AttriHuman-3D achieves strong disentanglement between different attributes, generates high-quality view-consistent 3D human avatars which allows fine-grained editing. From left to right, we show the generation results by editing different attributes, try-on results by modifying the selected attribute sets and generation results of view-consistent images.}
\end{figure*}

% As the increasing demand for digital content creation in various domains such as gaming, virtual reality, and e-commerce, the need for realistic and customizable 3D human avatars has grown significantly. In this context, building 3D-aware human avatar generation models from 2D image collections have recently attracted much more attention~\cite{bergman2022generative, hong2022eva3d, zhang2022avatargen, noguchi2022unsupervised}. Most of the recent researches explore 3D-aware human avatar generation as an extension of the 3D generative models for rigid objects~\cite{chan2022efficient, or2022stylesdf}~(e.g. natural objects, human faces and so on), which represents the objects explicitly or implicitly in the neural radiance fields. While achieving satisfying performance on generating 3D view-consistent human avatars, they failed to support user-interacted local editing, which is very important for downstream applications in the real-world scenarios. 

% human editing 
As the increasing demand for digital content creation in various domains such as gaming, virtual reality and e-commerce, the need for realistic and customizable 3D human avatars has grown significantly. Traditional 2D editable generation models control the generation style with the manipulation of high-level style latent codes~\cite{karras2019style, karras2021alias, karras2020analyzing}, which is ambiguous and hard to achieve precise local editing, limiting its application in the real-world scenario. 
Recently, several approaches have been proposed to enable user manipulation on the free-view generation results of 3D-aware GANs for rigid objects~\cite{sun2022fenerf, sun2022ide, ma2023semantic}~(e.g. human faces and cars). However, editable 3D-aware generation for human avatars could be more challenging due to its high variance in human geometry and appearance. 

Some existing editable 3D-aware GANs for human portraits
represent the overall regions of the generated objects with a single NeRF and perform local editing by asking users to manually edit the generated semantic masks~\cite{sun2022fenerf, sun2022ide}. Although being very efficient, these methods suffer from the high variance of human shapes, poses and outfits. Directly manipulating the semantic masks requires strong painting skills for the users and involve a lot of tedious manual works, making it hard to be applied in the real-world scenarios. One possible solution is to model and manipulate different semantic regions independently. CNeRF~\cite{ma2023semantic} proposes to model each semantic region with different generators. Although generating different semantic parts independently, training several deep generators in parallel will extremely increase the memory footprint, hindering their applications in real-world scenarios. Moreover, currently available human datasets are limited and almost all of them suffer from the style entanglement between different attributes, which means the style of one attribute may be highly influenced by the style of another attribute. For example, some pairs always show in the existing datasets: dresses always with high-heel shoes, t-shirts always with sports shoes. Men rarely with dresses. Such correlation may also be learned by the discriminator, leading to misleading punishment in the training process and artifacts in the editing stage.

% Attributes distribution between different genders could be very different. Men wearing dresses seldom appears in the real-world datasets. Sampling attribute sets in the training process may lead to misleading punishment from the discriminator and degrade the generation quality. Moreover, the shape of a certain hairstyle or cloth could be dependent on the body identity of different human, thus simply modeling all the attributes independently without any global-level style control may lead to artifacts in the test time.  %  The decomposition progress should be hierarchical. % $2)$ The sampling strategy for sampling different factors (e.g. hair, clothes and body shapes) to composite one human avatars could highly influence the training process. The frequency of different factors in the human datasets can be highly imbalanced, which may hinders the unsupervised training of 3D GANs. Moreover, the gender should also be considered in the sampling progress. For example, a "man" wearing a "dress"  seldom appears in the current available datasets, sampling this kind of factor combination may lead to misleading punishment from discriminator thus degrades the generation quality. Therefore a proper sampling strategy is also crucial to alleviate these issues. 

% represent in a 4D volume, 
To address these problems, we propose AttriHuman-3D, an editable 3D human generation model with attribute decomposition and indexing. Specifically, we propose to generate all attributes~(e.g. hair, clothes, main body and so on) in an overall feature space with six feature planes and then extract independent attribute-specific features with different attribute indexes. Different from CNeRF ~\cite{ma2023semantic} that generates each attribute with an independent generator, our method benefits from the "all-to-one" design, facilitating the information sharing across different attributes and achieving much higher computational efficiency.  Our method is inspired by the recently proposed tensor decomposition technologies in 4D dynamic NeRFs~\cite{cao2023hexplane, fridovich2023k}, we formulate the generation space as a 4D space-attribute field which could be decomposed into six feature planes and efficiently generate with a single 2D CNN-based feature generator. However, trivially extending the tensor decomposition technologies in space-time field into space-attribute domain leads to degraded generation results. Different from time dimension which is continuous and has clear definition, the attribute dimension could be disjoint and ambiguous, making it hard to define the index of different attributes. To address this problem, we further propose an implicit indexing module to learn the index of each human attribute in the attribute dimension. To enhance the disentanglement of different attributes, we further propose an orthogonal projection regularization to enforce the orthogonality of different indexes, which enables us to extract independent attribute features from the generated feature planes. % Moreover, we also propose an orthogonal projection regularization to enhance the disentanglement of learned indexes.  % In detail, the hybird explicit-implicit indexing method consists of a fixed explicit index and a learnable linear implicit index, where the fixed explicit index provides inductive bias to encourage the index to be uniformly distributed in the factor dimension while the learnable linear implicit index learns residuals to adjust the final index distribution. 

% Our framework is built on EG3D~\cite{chan2022efficient}, involving a StyleGAN2-based~\cite{karras2020analyzing} generator, an implicit attribute indexing module, a compositional neural volume renderer and a 2D CNN-based up-sampler. During rendering, we sample the selected attribute sets from the real-world dataset, extract the sampled features from the generated feature planes leveraging the implicit index module, then fuse and render all the extracted attributes with compositional volume rendering. % To address the imbalanced distribution of different factors in the dataset, we re-balance the distributions by sampling more factors with low-frequency of appearance. In addition, we found the distribution difference of factors between male and female highly influences the training process.
To address the implicit style entanglement between different attributes in the existing datasets, we 
introduce a hyper-latent training strategy which conditions the overall training progress with hyper-latent label to avoid the misleading punishment from the discriminator and an attribute-specific sampling strategy to split different attributes with pre-defined sampling bounding boxes to remove the influence of other attributes defined in different bounding boxes.% In this way, we avoid the misleading punishment from the discriminator to the generated images such as "a man wearing a dress" and the style correlation between different attributes, leading to better generation results. % We also involve a light-weight global generator to provide global style information to all sampled attributes, encouraging the final generated images to be more coherent with independent sampled attributes. % To solve this problem, we introduce a gender-wise strategy by conditioning the overall generation with the input gender option. In this way, we avoid sampling edge cases as "a man wearing dress" in the training, which stabilizes the training process and leads to much better generation results.

% In order to generated high-quality animatable 3D human avatars, we disentangle the pose and identities of the generated human avatars by shaping the overall space-attribute fields in the canonical pose space and find the deformation between canonical space and each observation posed space by sampling the pre-defined skinning weights on the SMPL human templates. To accommodate the non-rigid deformation between the observation and the canonical spaces, we involve a per-attribute non-rigid deformation network to predict the non-rigid deformation offsets for each attribute. 

Qualitative and quantitative experiments conducted on the fashion datasets demonstrate that our model provides a strong disentanglement between different attributes, allows fine-grained
image editing and generates high-quality view-consistent 3D human avatars. In summary, our main contributions can be listed as follows: 1) We propose a novel editable 3D human avatar generation model, which achieves fully disentangled control over the generated human avatars with attribute decomposition and indexing. 2) We propose a novel implicit indexing method with an orthogonal projection regularization to enhance the disentanglement of different attributes. 3) We introduce a hyper-latent training strategy and an attribute-specific sampling strategy to address the style entanglement between different attributes in the existing human datasets, leading to better editing performance.
\section{Related Work}

\begin{figure*}
  \centering
    \includegraphics[width=0.85\linewidth]{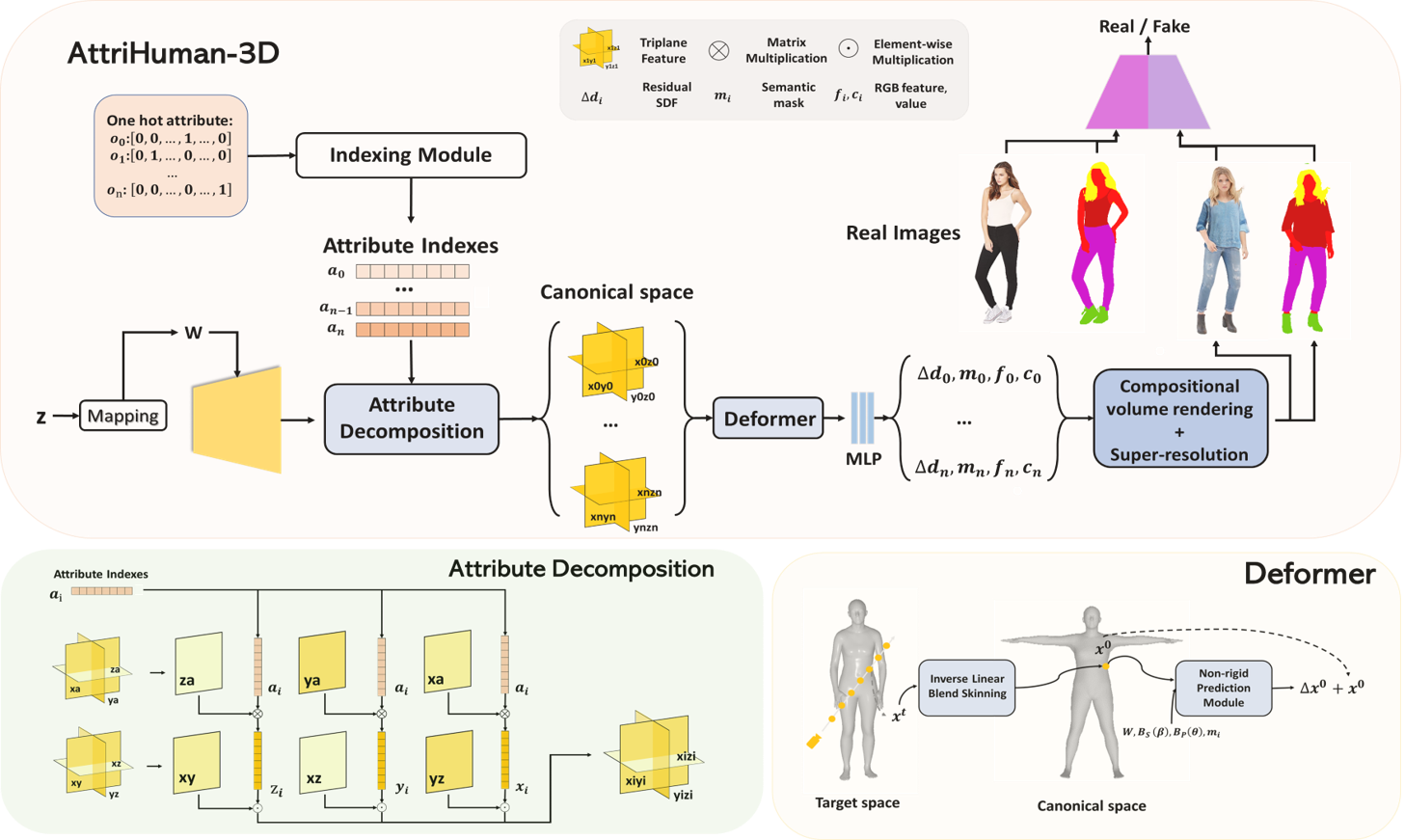}
  \caption{The overall framework of our model. We generate the decomposed feature plane with StyleGANv2-based generator and predict the indexes of each attribute with implicit indexing module. We model the deformation between canonical space and target space with deformer module and synthesis final image with compositional volume rendering and super-resolution module. Detailed structure of the attribute decompose module and deformer module is shown at the bottom, where $B_{s}(\beta)$, $B_{P}(\theta)$ represents the SMPL parameters randomly sampled from the dataset, n denotes the total number of selected attributes.}
      \label{pipeline}
\end{figure*}

\subsection{3D-Aware GANs}

Generative Adversarial Networks (GANs)~\cite{goodfellow2014generative} have achieved remarkable success in the field of 2D image synthesis~\cite{brock2019large, karras2018progressive, karras2019style, karras2020analyzing}. % Recent efforts have been made to extend these capabilities into 3D-aware image generation. Early methods build 3D GANs with meshes~\cite{liao2020towards, gao2022get3d}, point clouds~\cite{li2019pu} and voxel grids~\cite{hao2021gancraft} as representations. 
Recently, the implicit neural representation have been widely explored to be combined with GANs to learn 3D-aware image synthesis with only 2D images as supervision~\cite{niemeyer2021giraffe, or2022stylesdf,chan2022efficient,chan2021pi}.
Among them,  tend to have inferior generative power due to limitations of 2D images and inconsistent perspectives. Further, the Neural Radiance Fields (NeRF)\cite{mildenhall2020nerf} has become the most vital method for novel view synthesis by volumetric rendering techniques\cite{kajiya1984ray}, while using neural radiance and density fields to represent the generated 3D scene. 
Most of methods focus on single-view supervised 3D generation.
Particularly, GRAF~\cite{schwarz2020graf} and Pi-GAN~\cite{chan2021pi} propose to learn 3D-aware image and geometry generation with implicit neural radiance fields as generators. However, these methods are limited to synthesis only low-resolution images due to the large computational costs of the MLP querying in the volumne rendering process. To solve this problem, recent methods tend to adopt a two-stage generation process by firstly rendering low-resolution images and then passing it to 2D CNN-based decoders to generate images with higher resolution~\cite{niemeyer2021giraffe,chan2022efficient,or2022stylesdf}. %Among them, StyleSDF~\cite{or2022stylesdf} proposes a style-based 3D generator and represents the geometry of the generated objects with signed distance fields for better geometry modeling. EG3D~\cite{chan2022efficient} proposes a tri-plane 3D representation to improve the computational efficiency. 
Although providing high-quality generation of 3D-aware images, these methods fail to support user-interacted local editings. FENeRf and IDE-3d~\cite{sun2022fenerf, sun2022ide} further proposes to achieve local editing by involving semantic masks as intermediate representations and perform user-interacted editing via manually manipulating on the generated semantic masks. CNeRF~\cite{ma2023semantic} represents each semantic category with an independent generation network. In the task of 3D human avatar generation, most of the recent works generate the 3D human avatars with predefined parametic human templates~\cite{zhang2023avatargen, hong2022eva3d, noguchi2022unsupervised, bergman2022generative, zhang20223d}. Among them, EVA3D~\cite{hong2022eva3d} proposes a compositional human NeRF representation which divides the human body into different local parts and achieves high-resolution human image generation. However, all of them do not support user-interacted editing of generated human avatars. Some recent works~\cite{hong2022avatarclip,cao2023dreamavatar,patashnik2021styleclip,haque2023instruct} also explore using text-image embedding to edit the generation results. Although achieving good performance, these methods can not achieve fully disentangled control over different semantic parts and suffer from long optimization time.   % However, as they represent the generated instances with an overall neural radiance field, they fail to support fine-grained editing on the multi-layer instances, such as human avatars. CNeRF~\cite{ma2023semantic} represents each semantic category with an independent generation network. Although it fully disentangles each semantic regions, training several generators in parallel extremely increases the computational costs, limiting its application in the real-world

\subsection{Spatial decomposition}
NeRF proposes to use fully implicit networks to represent 3D density and color fields, which is slow in querying and computational infeasible to generate high-resolution images directly. A recent trend is to reduce the rendering time of NeRF with explicit geometric representations, including sparse 3D grids~\cite{yu2021plenoxels}, point clouds~\cite{xu2022point} and so on. However, there is a tradeoff between the rendering speed and the memory costs of different geometric representations. Although explicit methods reduce the optimization time, it requires more memory footprint. 

Hybrid representation~\cite{chan2022efficient, chen2022tensorf, muller2022instant, fridovich2023k, cao2023hexplane} has been proposed to improve the computational efficiency of NeRF by combining the explicit representations with implicit layers. In particular, %the explicit representations are first compressed by certain spatial decomposition methods and then decoded with small implicit networks. 
% Instant-NGP decompose the explicit 3D volumes with a multiresolution voxel grid, which is indexed with a hash function. 
TensoRF~\cite{chen2022tensorf} proposes to decompose the voxel tensor into feature planes and vectors, which greatly improves the computational efficiency. While all these methods focus on the decomposition of 3D feature volumes. Hexplane~\cite{cao2023hexplane} further extends it into 4D dynamic scenes. They decompose the 4D space-time volume into six feature planes spanning each pair of coordinate axes (e.g. $XY$, $ZT$).%  In this work, we adopt a similar decomposition method as Hexplane while extending it into a much more challenging task, which decomposes a static volume into different semantic regions.
\section{Method}

\subsection{Efficient 4D Space-Attribute Decomposition} \label{E4D}
% We propose a novel global-local hierarchical attribute decomposing method to fully decompose attributes in the generated 

% esides, the decomposition scheme also introduces implicit constraints on the
% representation since only low-rank tensors can be approximated by a small number of lower-dimensional components. Such a constraint can serve as an inherent regularization when the input observation is limited
%To fully disentangle different human attributes in the spatial position,
%we propose to represent t
we formulate the generated human avatars in a 4D space-attribute field (e.g. hair, clothes, body shapes and so on), which could be denoted as $(x,y,z,a) \in \mathbb{R}^4$  where $a \in \{clothes, hair, shoes ...\}$ denotes the attribute dimension. Given a selected attribute $attr_i$ and its attribute index $a_i = Index(attr_i)$, we are capable to extract the corresponding attribute features independently from the generated feature planes.  Recently, CNeRF~\cite{ma2023semantic} propose to generate the 4D volumes~$(x,y,z,a_i), i \in \{1,2,...,N\}$ with $N$ independent generators. However, training several independent generators in parallel will exponentially  increase the computational cost. Inspired by the tensor decomposition technologies in 4D dynamic NeRFs~\cite{cao2023hexplane}, we propose to decompose the 4D space-attribute field into six feature planes. Specifically, we decompose the 4D space-attribute volume $V\in \mathbb{R}^{XYZAF}$ into six planes : $P^{XY}_r$, $P^{YZ}_r$, $P^{XZ}_r$, $P^{XA}_r$, $P^{YA}_r$, $P^{ZA}_r$, and represent the space-attribute fields $D$ as:
\begin{equation}
\begin{aligned}
        D(x,y,z,a) = (P_{xy}^{XYR_1} \odot P_{za}^{ZAR_1})V^{R_1F} \\ +
    (P_{xz}^{XZR_2} \odot P_{ya}^{YAR_2})V^{R_2F} \\ + (P_{yz}^{YZR_3} \odot P_{xa}^{XAR_3})V^{R_3F},
\label{hexp}
\end{aligned}
\end{equation}
where $\odot$ denotes the element-wise product.  In our experiments, we find that relaxing Eq~\ref{hexp} by setting $V^{R_1F}$, $V^{R_2F}$ and $V^{R_3F}$ to be constant vectors leading to faster convergence without losing performance. Therefore, we set $V^{R_1F}$, $V^{R_2F}$ and $V^{R_3F}$ to be constant in our all experiments.

Leveraging the proposed tensor decomposition technology, we are capable to efficiently generate the decomposed feature planes with a single 2D CNN-based feature generator (like StyleGan2~\cite{karras2020analyzing}), which greatly reduces the computational cost as well as facilitates information sharing across different attributes and encourages consistency among the generated features of different attributes.

%  By utilizing and factorizing an explicit 4D tensor, our method enables both efficient reconstruction and compact representation of dynamic scenes.

% Since some planes rely only on spatial coordinates
% (e.g. XY ), by construction a HexPlane encourages sharing
% information across disjoint timesteps.

% In this way, we sample different we achieve fully disentanglement for different attributes with a single 4D fields. However, one chanllenge is that involving an additional attribute dimension will exponentially increase the memory footprint and computational cost. To solve this problem, we propose to adopt 4D tensor decomposition technologies which is recently proposed in dynamic NeRFs. 4D tensor decomposition technologies have been widely used to reduce the computational cost of dynamic 4D NeRFs: Cao proposes HexPlane, which decomposes the 4D dynamic scenes with six feature planes; Shao decomposes the 4D tensor hierarchically by projecting it first into three time-aware volumes and then nine compact feature planes.

\subsection{Implicit Attribute Indexing} \label{AI}
We propose implicit attribute indexing module to predict the indexes of different attributes. As the
attribute dimension could be disjoint and ambiguous,
trivially adopting the identical mapping indexing method as dynamic NeRFs~\cite{cao2023hexplane} leads to degraded generation results (Figure~\ref{abla_index}). To address this problem, we propose an implicit indexing module to learn the index of each predefined attribute in the attribute dimension.
In detail, the implicit indexing module $I$ is formulated as an MLP network with eight layers. The input for the indexing module is a one-hot label $o_i \in \mathbb{R}^N$, where $N$ is the number of total attributes. To stabilize the training, we normalize the predicted index before using it for attribute indexing. The total function of the implicit indexing module can be simply formulated as $
    a_i = norm(I(o_i))$, 
where $norm$ denotes the $L2$ normalization. The independent feature $D_i$ for attribute $i$ can then be extracted from the generated feature fields $D$ as:
\begin{equation}
\begin{aligned}
        D_i = (P_{xy}^{XYR_1} \odot (P_{za}^{ZAR_1} \times a_i^{AR_1} )) \\ +
    (P_{xz}^{XZR_2} \odot (P_{ya}^{YAR_2} \times a_i^{AR_2})) \\ + (P_{yz}^{YZR_3} \odot (P_{xa}^{XAR_3} \times a_i^{AR_3})).
\end{aligned}
\end{equation}
Note we omit $V^*$ here for simplicity. 
 During our experiments, we found that learning the indexing module in an unsupervised style fails to achieve strong disentanglement between different attributes, leading to collapsed geometry and chaotic textures in the editing stage. To address this problem, we further propose an orthogonal projection regularization to enforce the orthogonality between indexes for different attributes.

\noindent
\textbf{Orthogonal Projection Regularization.} Given a predicted index $a_i$, the orthogonal projection regularization could be defined as:
\begin{equation}
    L_{orth} = \sum_{i,j\in B \atop i\neq j}{|\langle a_i, a_j \rangle}|, \qquad \langle x_i, x_j \rangle = \frac{x_i \cdot x_j}{\Vert x_i \Vert_2 \cdot \Vert x_j \Vert_2},
\end{equation}
which involves no additional learnable parameters and can be applied directly on the mini-batch level for each iteration. In our experiments, we found that the orthogonal projection regularization is crucial to enforce the implicit indexing module to learn orthogonal attribute indices, facilitates the disentanglement between different attributes and leads to better editing results. 

%  While the orthogonal projection regularization is capable to enforce the orthogonality in the indexing level, the generated  

% it may fail to preserve the . To address this problem, we further extend it into cross-batch level with mix-attribute regularization. Given a generated human $H_1$ with sampled attributes $(a^{h1}_0, ..., a^{h1}_j), j<N$, where $N$ is number of total available attributes. We mix the sampled attribute 

% \textbf{Orthogonal Projection Regularization.} The orthogonal projection regularization is proposed to . 
% In our experiments, the orthogonal projection regularization is applied on the mini-batch level for each iteration.

% \\
% \textbf{Mix-attribute Regularization.}

% The indexing module is independent of the sampled latent code $\omega$, 

% decomp
% Extracting the feature from 4D fields is no trival as the feature dimension can be complex and no linear. Simply use a fix indexing methods may leads to degrade performance. To 
 % disjoint

% identity mapping  

\begin{figure*}
  \centering
    \includegraphics[width=\linewidth]{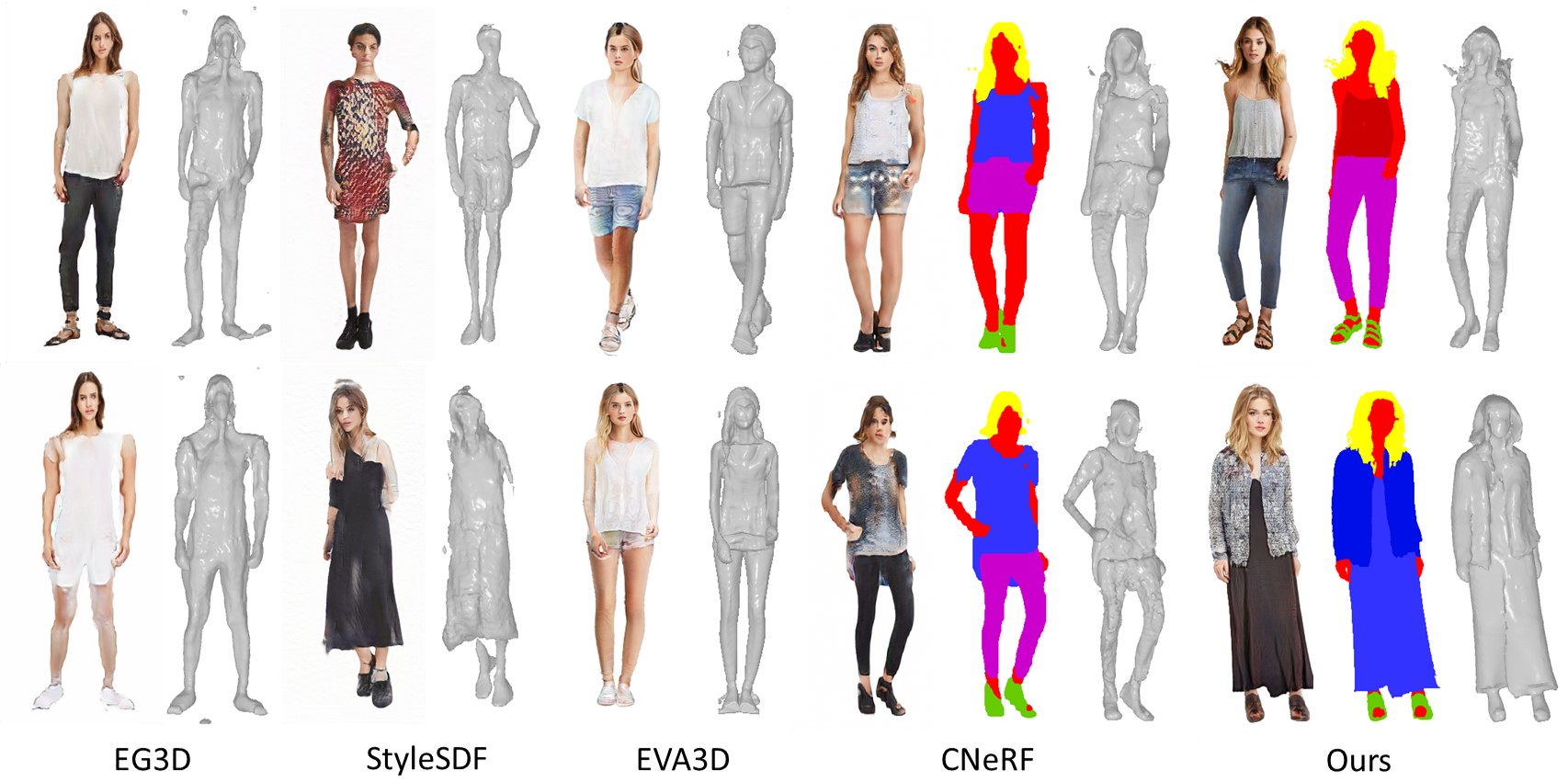}
\caption{Qualitative Comparisons of our methods with EG3D, StyleSDF, EVA3D, CNeRF. The RGB images, generated segmentation masks and 3D meshes demonstrated that our method achieves high-quality human avatar generation. Moreover, the main contribution of our model is to support interactive user editing, which is not supported by EG3D, StyleSDF and EVA3D.}
    \label{qua}
\end{figure*}

\subsection{Overall Framework}
\label{3DF}

Our framework is built on EG3D~\cite{chan2022efficient}, involving a StyleGAN2-based~\cite{karras2020analyzing} feature generator, an implicit attribute indexing module, a compositional volume renderer and a 2D CNN-based up-sampler. The overall framework of our model is shown in Figure~\ref{pipeline}. 

In detail, we first adopt a StyleGAN2-based generator to generate the overall feature planes. We then sample a set of attributes $O_i$ from the real-world dataset, where $O_i = (o_0, ..., o_k), k<N$, $N$ represents the total number of attribute categories predefined on the dataset. In our experiments, we set 11 attribute categories including the human body, hair, shirts, pants, skirts, dress and so on. During the generation process, each attribute is embedded as a one-hot label $o_i$ and is passed into the implicit indexing module to get the learned index in the attribute dimension. The features of each attribute are then extracted from the generated feature planes. Given a sampled point on a certain camera ray, we query its feature vectors by projecting the points onto the axis-aligned extrated feature planes and decode the residual signed distance values $\Delta d_i$, semantic masks $m_i$, RGB features and values $f_i, c_i$ with a multi-head MLP, which shares the weights across different attributes.

We disentangle the pose and shape of human by shaping the generated space-attribute fields in a common canonical space and find the deformation between the canonical space with each observation posed space leveraging the pre-defined skinning weights on the SMPL human templates.  In detail, we define the transformation
of a point $x^t$ from observation space $t$ to canonical space as:
\begin{equation}
\begin{aligned}
    \left[ x^0 \atop 1 \right] = \sum_{v_i \in \mathcal{N}(x)}{\frac{\omega_i}{\sum{\omega}}M(\beta_0, \theta_0)(M(\beta_t, \theta_t))^{-1}} \\ ,
    M(\beta, \theta) = \begin{bmatrix} I & B_{S}(\beta) + B_{P}(\theta) \\ 0^T & 1 \end{bmatrix},
    \label{o2ct}
\end{aligned}
\end{equation}
where $\mathcal{N}$ represents the nearest $k$ points that are found among the vertices of SMPL posed mesh $M(\beta_t, \theta_t)$, $w_i = 1 / \Vert x^t - v_j \Vert$ is the transformation weight, $B_{S}(\beta)$ and $B_{P}(\theta)$ denote the shape/pose blend shape for vertexes on the SMPL mesh, respectively.  Similar to AvatarGen~\cite{zhang2023avatargen}, we further involve a per-attribute non-rigid deformation network, which predicts the non-rigid offsets $\Delta x^t$, to accommodate the non-rigid deformation between the observation and the canonical spaces. Denote the transformation defined in E.q.~\ref{o2ct} as $T$, the final deformation can be formulated as $x^0 = T(x^t) + \Delta x^t$, $\Delta x^t =  MLP(embed(x^0), B_{S}(\beta),B_{P}(\theta), m_i)$, where $m_i$ represents the semantic mask for attribute $i$.

As we formulate all the generation in the canonical space, it's easy for us to conduct editing by directly exchanging the feature planes. To edit certain attribute $o_i$, we could change the generated feature planes of $o_i$ with another ones generated from new sampled latent codes. In this way, we support precise editing on certain selected attributes while keeping others fixed. We show more details in the supplemental materials.

\subsection{Compositional Volume rendering}
\label{CAr}
We fuse and render the output of all attributes with compositional volume rendering. Specifically, we first fuse the outputs of all selected attributes with the predicted semantic masks. Given a sampled points $x$ in attribute $i$ with predicted residual signed distance values $\Delta d_i$, semantic mask $m_i$, RGB features and values $f_i$, and $c_i$, the fusion step can be formulated as:

% \begin{equation}
% m'(x) = \frac{exp(m_i(x)}{\sum_j^N{exp(m_j(x)}}
% \end{equation}
\begin{equation}
d(x) = d_t(x)+ \sum_i^N{m'_i(x) \cdot \Delta d_i(x)},
\end{equation}
\begin{equation}
f(x) = \sum_i^N{m'_i(x) \cdot f_i(x)}, \qquad c(x) = \sum_i^N{m'_i(x) \cdot c_i(x)},
\end{equation}
where $m'_i(x) = \frac{exp(m_i(x))}{\sum_j^N{exp(m_j(x))}}$ denotes the softmax operation, $d_t$ represents the template SDF queried from the SMPL template, N denotes the total number of semantic categories. %  We then transform the signed distance value d into the density value $\sigma$ as $\sigma(x) = \frac{1}{\sigma} \cdot Sigmoid(\frac{-d(x)}{\alpha})$. For each ray $\mathrm{r}(t) = \mathrm{o} + t\mathrm{v}$, we query the sampled points and integrate all the values with volume rendering equations:
% \begin{equation}
%     I(R) = \sum^N_{i=1}\left( \prod_{j=1}^{i-1}e^{-\sigma_j \cdot \delta_j }\right)  \cdot (1 - e^{-\sigma_i \cdot \delta_i}) \cdot f_i,
% \end{equation}
% where $\delta = \Vert x_i - x_{i-1} \Vert $ and $N$ is the number of sampled points along each ray.  
We then adopt volume rendering~\cite{mildenhall2020nerf} to render the low-resolution RGB image and semantic mask. The rendered RGB feature along with the segmentation masks are then fed into
the super-resolution module to generate the final high-resolution images.

\subsection{Training}
\label{train}

\textbf{Hyper-latent Training Strategy.} 
During our experiments, we found that the style entanglement existing in the current human datasets~\cite{liu2016deepfashion}. One of the most significant problems is that almost all datasets have few or no images containing men wearing dresses. However, as attribute is defined according to semantic areas without definition of human genders. Problems come when we generate a body area of men with attributes of dress. In this condition, the generated image would become out-of-distribution to the discriminator and it would give misleading punishment. As a result, the feature plane of attribute dress would try to influence the feature of attribute human body, making it to look like women~(Figure~\ref{abla_asp}), leading to slower convergency and artifacts in the editing stage. To address this problem, we condition the linear mapping module of StyleGAN~\cite{fu2022stylegan} with hyper-label to control the gender of the generated human bodies and sample reasonable attribute sets according to different genders. In this way, we avoid the misleading punishment from the discriminator and achieves better generation results.

\noindent
\textbf{Attribute-specific Sampling Strategy.} While hyper-latent training strategy is adopted to address the entanglement between attributes with large spatial overlap, like body and dress. We propose attribute-specific sampling strategy to address the entanglement between attributes with little overlap areas. For example, the style entanglement of dresses always with high-heel shoes, T-shirts always with snipers. %As a result, the style of generated shoes would be dependent on the appearance of the dresses, which leads to artifacts in the editing stage. To address this problem and further improve the computational efficiency, 
In detail, we define
attribute-specific bounding boxes for each attribute and sample corresponding attribute feature only inside the bounding box. In this way, we limit the influence of each attribute strictly inside the bounding box, enhance the disentanglement between attributes with little or no overlaps and further improve the computational efficiency by sampling much fewer points for each attribute. We present more details of the pre-defined bounding boxes in the supplementary material.

% \textbf{Discriminator.} To model the joint distribution of real RGB images and semantic masks, We adopt a dual-branch discriminator $D_{dual}$, which takes both RGB images and segmentation masks as input. We feed the SMPL parameters~\cite{loper2015smpl} and gender label into the discriminators as conditions to further stabilize the training process. Moreover, to enhance the generation quality of the human face, we further involve a face discriminator $D_{face}$. In detail, we crop the head regions from the generated high-resolution fake images and feed it into another shallow face discriminator
% for comparison with the real ones. 

\noindent
\textbf{Loss.} We use the non-saturating GAN loss $L_{gan}$ with R1 regularization $L_{Greg}$. We regularize the learned SDFs by pushing the derivation of delta SDF values to be zero: $L_{eik} = \sum_{x}{(\Vert \nabla d_i (x) \Vert) - 1)^2}$, and further prevent false or non-visible surfaces with $L_{surf} = \sum_{x}{exp(-100\vert d_i(x) \vert)}$. Moveover, to constrain the generated residual signed distance fields to be consistent with the SMPL template mesh, we guide the predicted residual signed distance fields with a minimum regularization: $L_{rsdf}=\sum_{x}(\Vert \Delta d(x) \Vert)$. To prevent the learned non-rigid deformation from collapsing, we add a non-rigid regularization to regularize the learned non-rigid deformation to be small: $L_{nonrig} = \sum_{x}(\Vert \Delta x^t \Vert)$.
The overall loss function is formulated as:
\begin{equation}
\begin{aligned}
    L_{overall} = L_{gan} + \lambda_{Greg} L_{Greg} + \lambda_{eik} L_{eik} + \lambda_{surf} L_{surf} \\+ \lambda_{rsdf} L_{rsdf} + \lambda_{nonrig} L_{nonrig} + \lambda_{orth} L_{orth},
\end{aligned}
\end{equation}
where $\lambda_*$ denotes the weights of each loss item. 
\section{Experiments}
\textbf{Datasets.}
We train and evaluate our model mainly on the Deepfashion dataset~\cite{liu2016deepfashion}, which contains single-view human images with different clothes and the corresponding semantic masks.  We select 11903 images that contain the full human bodies, crop and resize all the images to coarsely align the human bodies. % We then apply an off-the-shelf pose estimator~\cite{pavlakos2019expressive} to  estimate SMPL parameters and camera parameters of each image. 
All
images are resized to $512 \times 512 $ for training. We use an off-the-shelf pose estimator~\cite{pavlakos2019expressive} to  estimate SMPL parameters and camera parameters of each image in the dataset. For each human image, we convert its corresponding semantic masks into attribute sets by checking whether the semantic region is above zero.  we adopt 11 attributes for training, including Outer, Top, Skirts, Dress, Pants, Rompers, Hats, Glasses, Body, Shoes and Haircut.% See more implementation details in the supplemental materials.% , which could be easily implemented with the unique function in the Pytorch~\cite{pytorch2018pytorch}. We randomly sample different the SMPL paramenters, camera parameters and attribute sets from the dataset for training.

% We use an off-the-shelf pose estimator~\cite{pavlakos2019expressive} to  estimate SMPL parameters and camera parameters of each image in the dataset. For each human image, we convert its corresponding semantic masks into attribute sets by checking whether the semantic region is above zero, which could be easily implemented with the unique function in the Pytorch~\cite{pytorch2018pytorch}. We randomly sample different the SMPL parameters, camera parameters and attribute sets from the dataset for training. For the Deepfashion dataset~\cite{liu2016deepfashion}, we adopt 11 attributes for training, including Outer, Top, Skirts, Dress, Pants, Rompers, Hats, Glasses, Body, Shoes and Haircut.

\textbf{Implementation Details.}
 Our generator is built on the top of StyleGAN2~\cite{karras2020analyzing}, similar as EG3D~\cite{chan2022efficient}. The learning rates of generator and discriminator are 0.0025 and 0.002, respectively. The gamma value of RGB image and semantic masks are set to 10 and 100 to prevent the semantic masks from dominating the backward gradient thus degrading the quality of generated images. Our model is trained in two stages by gradually increasing the neural rendering resolution from $64$ to $128$.  
 
% \textbf{Implementation Details.}
% We build our framework on the top of EG3D~\cite{chan2022efficient}, and further extend it into multi-head by adopting separate linear layers to generate the geometry features and texture features from the shared feature map of StyleGAN2~\cite{karras2020analyzing} backbone. The learning rates of generator and discriminator are 0.0025 and 0.002, respectively. The gamma value of RGB image and semantic masks are set to 10 and 100 to prevent the semantic masks from dominating the backward gradient thus degrading the quality of generated images. Our model is trained in two stages by gradually increasing the neural rendering resolution from $64$ to $128$.  

\subsection{Comparisons.}
\textbf{Baselines.} We compare our AttriHuman-3D model with several state-of-the-art 3D-aware GANs, including EG3D~\cite{chan2022efficient}, StyleSDF~\cite{or2022stylesdf}, CNeRF~\cite{ma2023semantic} and EVA3D~\cite{hong2022eva3d}. EG3D, STyleSDF and CNeRF are designed for 3D generation of rigid / half-rigid objects such as human faces. Among them, CNeRF is also capable to support user-interacted editing for the generation results, which is similar to our task. EVA3D is also designed for human avatar generation, but it does not support interactive editing, which is the main contribution of our model.

% As recent researches have proofed that, due to the high variance of human poses, directly applying 3D aware GANs on the 3D human generation tasks leads to heavily mode collapsion and distorted generation results~\cite{zhang2023avatargen, hong2022eva3d}. We thus build a higher baseline by adapting these models with a similar deformation module as ours, which greatly reduces the training difficulty. 

% Figure 1. Edit results'
% Figure 3. Qualitative comparisons
% Figure 4. EG3d / CNERF / Ours
% Figure 5. Inverse CNERF / OURs
% Figure 6. Ablation: Gender / orthogonal / fix / mix

% Table 1. Quantititative 
% Table 2. Ablation

\begin{table}
        \scalebox{0.75}{
        \begin{tabular}{lccccc}
            \hline
            Methods & $\text{FID}\downarrow$  & $\text{FID}_{edit}\downarrow$ & PCK$\uparrow$ & Depth$\downarrow$ & \text{E} \\
            \hline\hline
            EG3D~\cite{chan2022efficient} & 25.90 & - & 79.63 & 0.0331 & $\times$ \\
            StyleSDF~\cite{or2022stylesdf} & 26.32 & - & 75.13 & 0.0376 & $\times$ \\
            CNeRF~\cite{ma2023semantic} & 27.69 & 30.31 & 71.56 & 0.0421 & \checkmark \\
            EVA3D~\cite{hong2022eva3d} & 15.91 & - & 87.50 & 0.0272 & $\times$ \\
            \hline
            fixed indexing & 21.20 & 23.58 & 88.41 & 0.0357 & \checkmark \\
    	w/o OPR & 19.81 & 22.31 & 88.72 & 0.0322 & \checkmark \\
		w/o HL & 17.24 & 18.11 & 89.63 & 0.0311 & \checkmark \\
		w/o ASP & 17.02 & 17.32 & 89.97 & 0.0305 & \checkmark \\
            Ours(full) &  16.85 & 17.43 & 89.91 & 0.0302 & \checkmark \\
            \hline
        \end{tabular}}
    \caption{Quantitative comparisons of our method with other methods, where OPR denotes orthogonal projection regularization, GS denotes Hyper-Latent training strategy and ASP denotes Attribute-specific Sampling strategy. E represents editable ability.}
    \label{quan}
\end{table}

 \begin{table}
 \begin{center}
        \scalebox{.85}{
        \begin{tabular}{lccc}
            \hline
            Methods & Memory & Parameters & Time(s)\\
            \hline\hline
            EG3D & 6G & 31M & 0.06 \\
            StyleSDF & 6G & - & 0.13 \\
            CNeRF & 16G & - & 0.23  \\
            EVA3D & 7G & - & 0.16  \\
            11G-base & Out of Mem & 311M & -\\
            Ours(full) &  9G & 59M & 0.09  \\
            \hline
        \end{tabular}}
 \end{center}
 \caption{Efficiency comparisons of our method with other methods. Our model is much more efficient compared to CNeRF and 11G-base model without using the proposed tensor decomposition technique. }
    \label{eff}
\end{table}

\noindent
\textbf{Quantitative Evaluations.}
The quantitative comparisons between our methods with baselines are shown in Table~\ref{quan}. In detail, we adopt three commonly
used image quality evaluation metrics including Frechet Inception Distance (FID)~\cite{heusel2017gans} to evaluate the visual quality and diversity of the rendered images, Percentage of Correct Keypoints (PCK)~\cite{bergman2022generative} to evaluate the effectiveness of the pose controllability. Pseudo Depth~\cite{ranftl2020towards} to evaluate the consistency between the generated geometry and RGB images. %We use $50K$ generated to compute the FID and KID metrics and use $5K$ images to compute PCK and depth metrics. 
As shown in Table~\ref{quan}, our model achieves comparable
results to the SOTA method EVA3D, while capable to support user-interacted editing. We also compare the FID of the edited images ($\text{FID}_{edit}$) with CNeRF and the result demonstrates our model achieves better editing results with a little drop in the FID for the edited images. Moreover, we also compare the computational efficiency of our model with others in Table~\ref{eff}, which shows our model only involves limited extra parameters to achieve editing capability. Benefiting from the proposed tensor decomposition technology our model is much higher efficient compared to CNeRF and 11G-base model which directly uses $11$ generators for the generation.

\begin{figure}
  \centering
\includegraphics[width=0.7\linewidth]{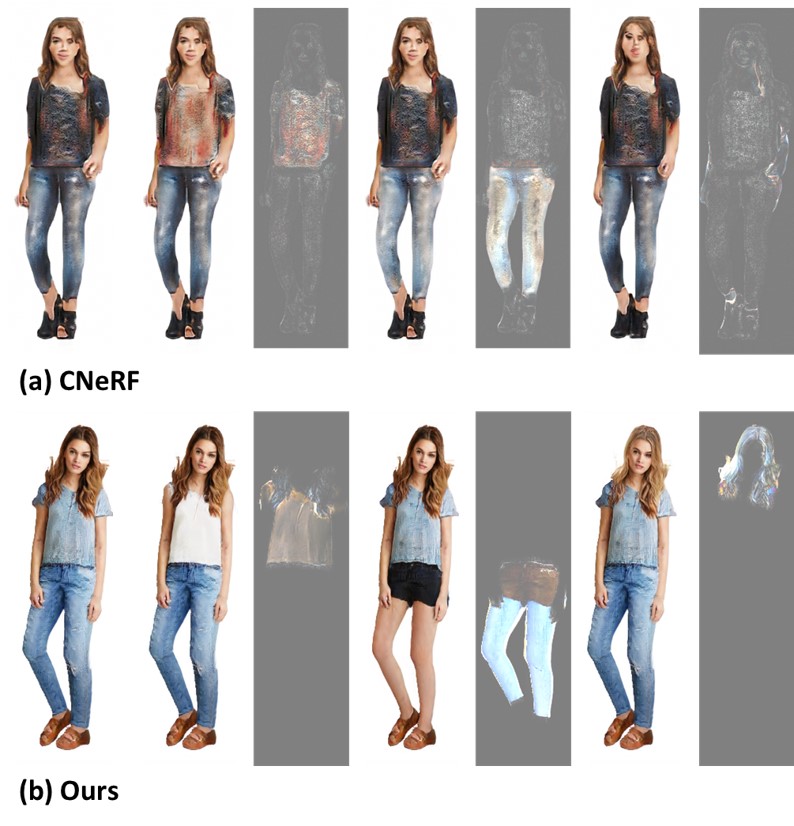}
\caption{ Qualitative comparisons of the editing results between our methods and CNeRF. From left to right we show the editing RGB and residual of changing the Top, Pants and Haircut. Benefit from our hyper-latent and 
 attribute-specific training strategy, our model achieves better disentanglement and more precise control over the target semantic region compared to CNeRF.}
    \label{cedit}
\end{figure}

\noindent
\textbf{Qualitative Evaluations.}
Figure~\ref{qua} shows the qualitative comparisons of our methods with other baselines. Specifically, we show the rendered images, segmentation masks and the corresponding meshes. 3D GANs designed for rigid/half-rigid objects including EG3D, StyleSDF and CNeRF fail to learn high-quality generation of human avatars due to the high variance of human poses and appearance. While EVA3D generates realistic 3D human avatars, it fails to support user-interacted editing. Our model achieves comparable generation results to EVA3D while capable to support semantic part-level editing that other methods do not support. We also compare the editing results of our model with CNeRF. As shown in Figure~\ref{cedit}, benefit from our hyper-latent and 
 attribute-specific training strategy, our model achieves better disentanglement and more precise control over the target semantic region compared to CNeRF.  % Without explicit human and deformation modeling, the generated shapes tends to be distorted with chaotic textures. Moreover, the boundary between clothes and human body is ambitious. As it models all the attributes in an overall neural radiance fields, it is hard to for it learn the semantic meaning of different regions. Although CNeRF is also capable for local editing of the generated 3D humans, it fails to model all the semantic categories correctly and suffers low rendering quality with undesired geometry. As shown in Fig~\ref{abla}, some face regions are missing and the generated clothes are noisy.  Compared to them, AttriHuman-3D not only generates high-quality 3D human avatars with vivid geometry and realistic textures, but also provides a strong disentanglement between different attributes, achieves fine-grained 3D-consistent human attribute editing.

\subsection{Ablations Studies.}

\begin{figure}
  \centering
\includegraphics[width=0.9\linewidth]{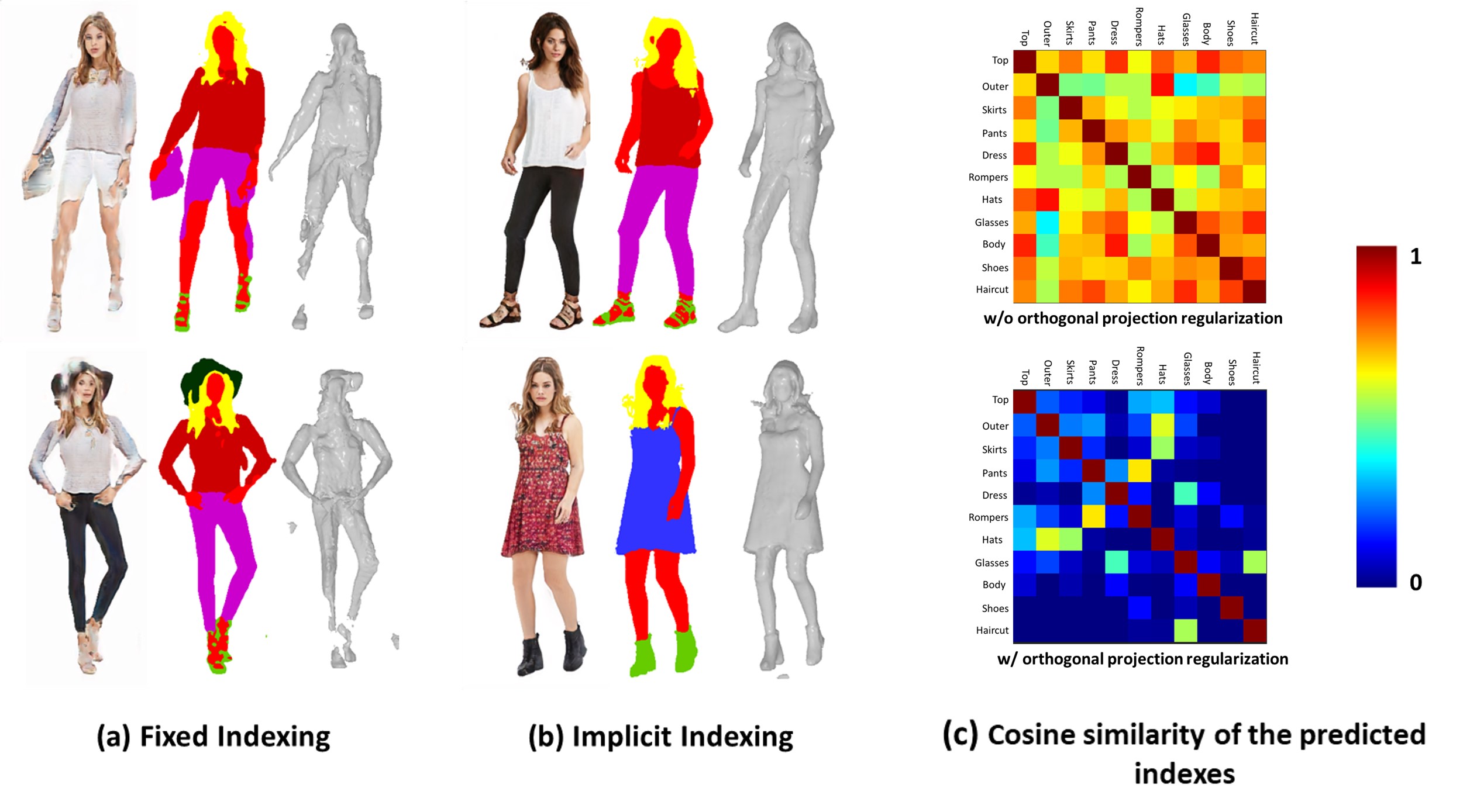}
\caption{Ablation of the implicit indexing module. Figure (a) and (b) compare the ablation results of our implicit mapping module with fixed identical mapping as dynamic NeRFs~\cite{cao2023hexplane}. Figure (c) shows cosine similarity of the predicted indexes with or without the proposed orthogonal projection regularization. }
    \label{abla_index}
\end{figure}

\begin{figure}
  \centering
\includegraphics[width=0.9\linewidth]{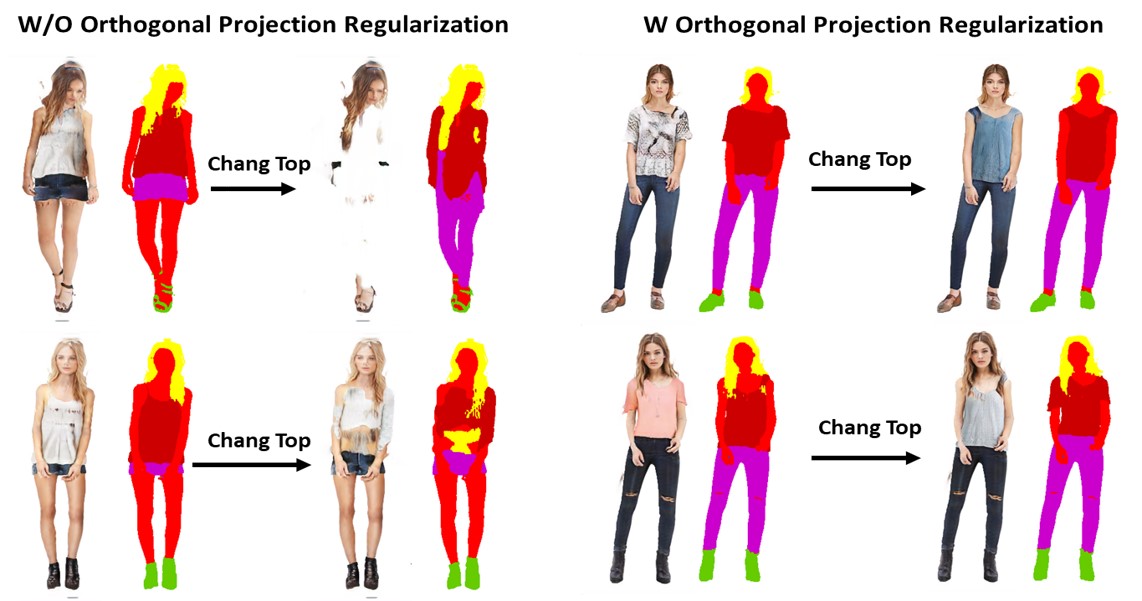}
\caption{Ablation of the orthogonal projection regularization. we achieve better editing results with the proposed OPR loss.}
    \label{abla_opr}
\end{figure}

\begin{figure}
  \centering
\includegraphics[width=0.9\linewidth]{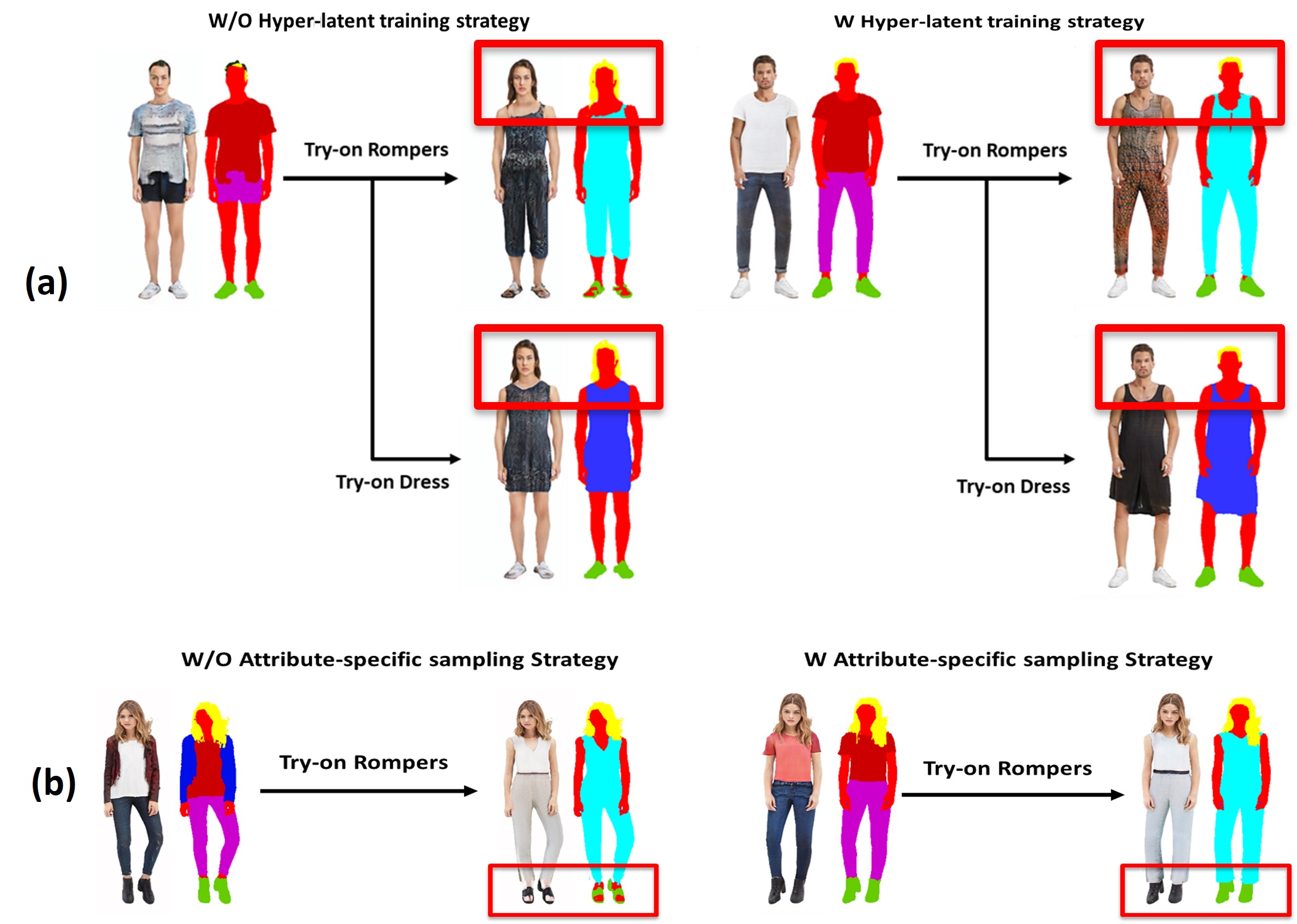}
\caption{Ablation of the Hyper-latent training strategy and Attribute-specific Sampling training strategy.}
    \label{abla_asp}
\end{figure}

\noindent
\textbf{Implicit Indexing Module.}
Trivially adopting the fixed identical indexing method in dynamic NeRFs~\cite{cao2023hexplane} leads to degraded generation results as shown in Figure~\ref{abla_index}.
In our implicit indexing module, the proposed orthogonal projection regularization serves an important role in encouraging the predicted implicit attribute indexes to be orthogonal to each other, facilitating the disentanglement between different attributes. As shown in Figure~\ref{abla_index}(c) , Figure~\ref{abla_opr}, without the orthogonal projection regularization, the learned attribute indexes tend be entangled with each other, leading to collapsed results in the editing stage. When we remove the original feature of the top clothes and change it into another one, the network fails to generate satisfactory edited images. We also conduct quantitative comparisons in Table~\ref{quan}. Removing the implicit indexing module or orthogonal projection regularization leads to a significant performance drop in all metrics, which indicates the effectiveness of the proposed implicit module and orthogonal projection regularization.
% We adopt Mix-Attribute Regularization to further constrain the generator to generate fully disentangled and independent features. Although we achieve strong disentanglement in the indexing level with the proposed orthogonal projection regularization, the generated features of different attributes may still be interrelated. This happens as human could be seen as multi-layer objects but only the out-most layer in the generated humans will be rendered and punished by the generator. Thus the generator tends to generate entangled attributes which forms a human avatar only in the outer-most layer.  However, the inner layer may contain artifacts or collapsed structures. As shown in left-bottom of Figure~\ref{abla} (a), when we try to edit the top clothes on the original generated human, the inner layer of the original human become visible and shows chaotic textures and collapsed geometry. Leveraging the proposed mix-attribute regularization,  we give a strong constrain on the generator not only in the outer-most layer but also in the inner layers, facilitate it to generate reasonable fully-disentangled features. Quantitative and Qualitative results in Table~\ref{quan} and Figure~\ref{abla} demonstrates our proposed mix-attribute regularization leads to better generation results.
% \textbf{Gender-wise Training Strategy.}

\noindent
\textbf{Training Strategies.} During training, we adopt the hyper-latent training strategy and attribute-specific sampling strategy to address the implicit style entanglement between different attributes in the existing dataset. As shown in Figure~\ref{abla_asp} (a), without the hyper-latent training strategy, the discriminator tend to give misleading penalty to push the attribute features of dresses or rompers to influence the body attribute and try to change the original appearance of the generated bodies to make them visually more suitable for a female subject. With the proposed hyper-latent training strategy, we avoid the misleading punishment from the discriminator and achieve more accurate editing results without changing the appearance of the generated human. The Attribute-specific sampling strategy enforces the style disentanglement between attributes with little or no overlay such as dress and shoes. As shown in Figure~\ref{abla_asp}, with attribute-specific sampling strategy, we avoid the style influence from rompers on the shoes, and achieve more precise editing results. Quantitative results shown in Table~\ref{quan} also indicate that the proposed training strategies lead to better generation results.

\begin{figure}
  \centering
    \includegraphics[width=0.95\linewidth]{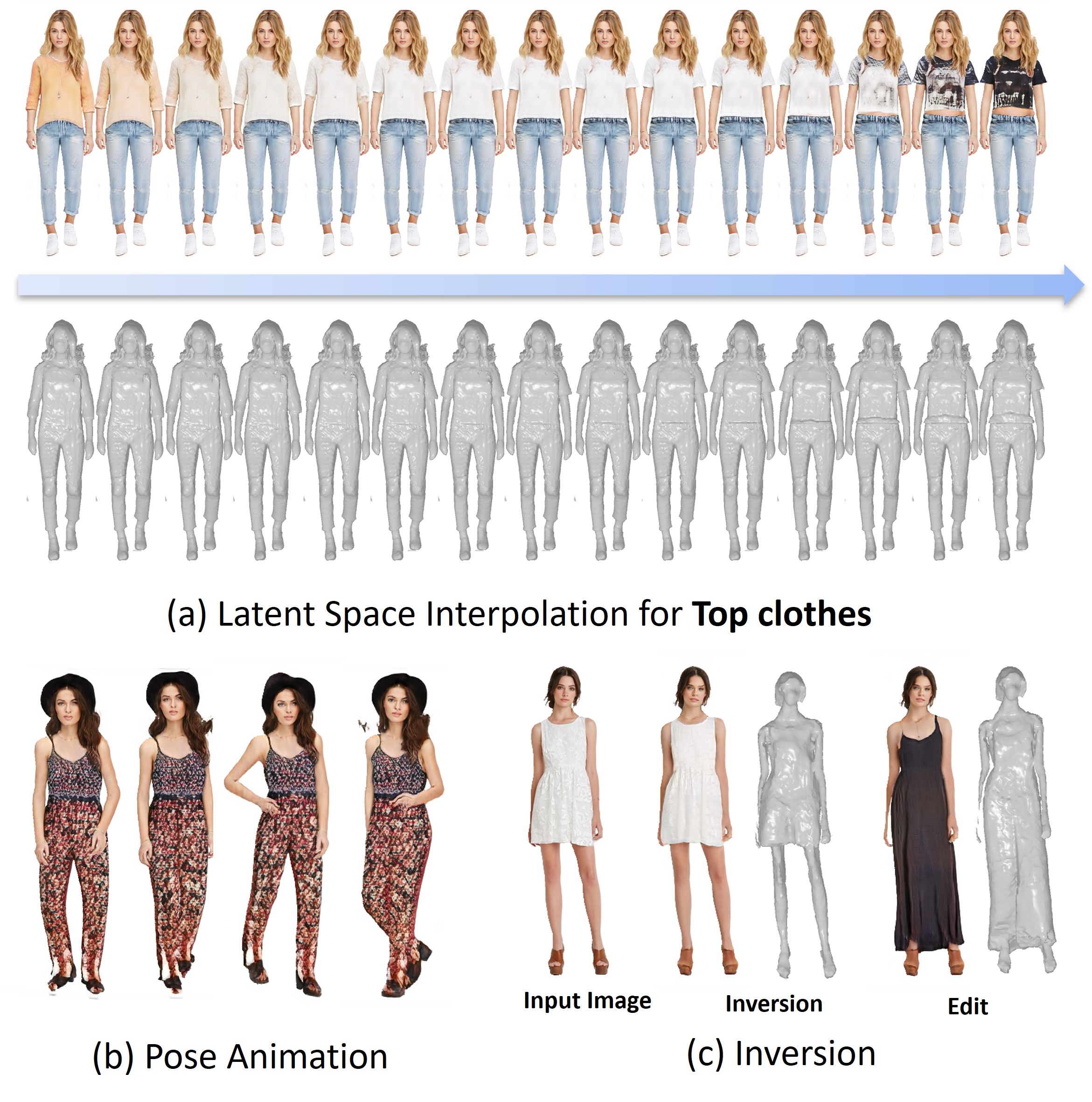}
    \label{app}
\caption{Application of AttriHuman-3D. (a) Interpolation on the latent code of Top Clothes gives smooth transformations between two samples. (b) Pose animation results in a generated avatar. (c) Inversion of real image}
\end{figure}
\section{Conclusion}
In this paper, we propose AttriHuman-3D, a novel editable 3D human avatar generation model.Our method
allows users to interactively edit selected attributes in the
generated 3D human avatars while keeping others fixed.
Both qualitative and quantitative experiments demonstrate
that our model provides a strong disentanglement between
different attributes, allows fine-grained image editing and
generates high-quality 3D human avatars
{
    \small
    \bibliographystyle{ieeenat_fullname}
    \bibliography{main}
}


\end{document}